# Protecting the Neural Networks against FGSM Attack Using Machine Unlearning

Amir Hossein Khorasani [1]     Ali Jahanian [2]     Maryam Rastgarpour [1]

[1] Department of Computer Engineering, Faculty of Engineering, Saveh Branch, Islamic Azad University, Saveh, Iran

[2] Faculty of Computer Science and Engineering, Shahid Beheshti University, Tehran, Iran

**Abstract:** Machine learning is a powerful tool for building predictive models. However, it is vulnerable to adversarial attacks. Fast Gradient Sign Method (FGSM) attacks are a common type of adversarial attack that adds small perturbations to input data to trick a model into misclassifying it. In response to these attacks, researchers have developed methods for "unlearning" these attacks, which involves retraining a model on the original data without the added perturbations. Machine unlearning is a technique that tries to "forget" specific data points from the training dataset, to improve the robustness of a machine learning model against adversarial attacks like FGSM.
In this paper, we focus on applying unlearning techniques to the LeNet neural network, a popular architecture for image classification. We evaluate the efficacy of unlearning FGSM attacks on the LeNet network and find that it can significantly improve its robustness against these types of attacks.

## 1  Introduction

Artificial Neural Networks (ANN) are a form of artificial intelligence that is designed to mimic the way the human brain works [1]. ANNs consist of interconnected nodes that are capable of processing and transmitting information, similar to the brains of creatures. Neural networks are used in a wide variety of applications, including image recognition, natural language processing, and predictive analytics [2].

One of the main advantages of neural networks is their ability to learn from data. They can be trained on large datasets to identify patterns and make predictions, which makes them particularly useful for tasks like image classification or speech recognition. Neural networks are also highly parallel, which means that they can process information much faster than traditional computers [3]. Another advantage of neural networks is their ability to generalize. Once they have been trained on a dataset, they can be used to make predictions on new data that they have never seen before. This makes them particularly useful for applications like predictive maintenance, where the goal is to identify potential problems before they occur. Overall, neural networks are a powerful tool for solving complex problems and making predictions based on large datasets [4].

One of the main concerns about neural networks is that they can be vulnerable to attacks. Several types of attacks can be launched against neural networks, including adversarial attacks, poisoning attacks, and backdoor attacks. These attacks are as follows [5]:

**Adversarial attacks** are designed to trick a neural network into misclassifying data. These attacks work by adding small, carefully crafted perturbations to the input data that are imperceptible to humans but can cause the neural network to make incorrect predictions.

**Poisoning attacks** involve manipulating the training data used to train the neural network. The goal of this type of attack is to introduce subtle changes to the training data that can cause the neural network to behave maliciously.

**Backdoor attacks** involve inserting a backdoor into the neural network during the training process. This backdoor can then be triggered by specific inputs, allowing an attacker to take control of the neural network.

Researchers are developing new techniques to make neural networks more robust and secure to mitigate these types of attacks. These contributions include techniques for detecting and removing adversarial examples, methods for detecting poisoning attacks, and ways to prevent backdoor attacks [6].

Overall, while neural networks are vulnerable to attacks, researchers are actively working to develop new techniques to make them more secure and resilient to these types of threats.

In this paper, we enhance the neural network's resistance against FGSM attacks through the machine unlearning implementation. Many deep learning models, it is susceptible to adversarial attacks such as FGSM. To protect against FGSM attacks, machine unlearning can be implemented. This involves periodically retraining the model on a dataset that includes adversarial examples. This process essentially "unlearns" the incorrect patterns that were learned by the model during the adversarial attack, making it more robust to future attacks.

This paper is organized as follows. section 2, we will examine the structure of neural networks. In section 3, we will introduce and examine machine unlearning and how it protects neural networks in our proposed method. In section 4, our proposed method explains the protection of neural networks against FGSM attacks and presents experimental results. Finally, we evaluate the obtained results.

### 1.1 Related work

A considerable amount of work has already been done in the area of neural network insecurities. In this section, we will briefly review and classify the literature. Several recent studies have explored the security vulnerabilities of neural networks. In particular, researchers have focused on the following topics:

**Adversarial attacks:** Goodfellow et al. (2014) [7] demonstrated that neural networks are vulnerable to adversarial attacks, in which small modifications to input data can cause the network to misclassify the data. This work has been extended by some subsequent papers, including Papernot et al. (2016) [8] and Carlini and Wagner (2017) [9].

**Backdoor attacks:** Gu et al. (2017) [10] introduced the concept of "backdoor attacks," in which an attacker can subtly modify the training data to introduce a flaw that can be exploited later. This work has been extended by Bhagoji et al. (2018) [11] and Liu et al. (2018) [12].

**Model stealing:** Tramer et al. (2016) [13] showed that it is possible for an attacker to "steal" a neural network model by querying the model and using the responses to reverse-engineer the network's architecture. This work has been extended by Juuti et al. (2018) [14] and Orekondy et al. (2019) [15].

By reviewing the work that has been done in this area, we can better understand the challenges and opportunities for improving the security of neural networks. In this article, we focus on increasing the security of neural networks against adversarial examples. Neural networks are susceptible to adversarial examples, which are carefully crafted inputs that cause the network to make the wrong prediction. This can have serious consequences in applications such as autonomous driving and medical diagnosis.

One approach by Aleksander Madry et al. (2017) [16], to increasing neural network security against adversarial examples is through adversarial training, where the network is trained on both clean and adversarial examples. This can increase the network's robustness to adversarial attacks.

Another approach by Nicolas Papernot et al. (2016) [17] is defensive distillation, which involves training a second network to mimic the behavior of the first network. The second network is then used to make predictions, making it more difficult for attackers to craft adversarial examples.

Other techniques include randomized smoothing by Cohen et al. (2019) [18], where the output of the network is perturbed with random noise to make it more difficult for attackers to craft adversarial examples, and feature squeezing by Xu et al. (2017) [19], where the number of colors or bits per pixel in the input image is reduced to make it more difficult to exploit small perturbations.

In past works, it was tried to make it more difficult to create adversarial samples, but in the method we present, the attempt is to remove the adversarial samples from the neural network, and thus the accuracy of the desired model can be significantly improved.

In this paper, we propose protecting the neural networks against FGSM attacks using machine unlearning. The Fast Gradient Sign Method (FGSM) attack is a popular adversarial attack technique used to generate adversarial examples to fool machine learning models. The idea behind the FGSM attack is to take the sign of the gradient of the loss function concerning the input and use it to perturb the input by a small amount in the direction that maximizes the loss.

## 2 The Structure of Neural Networks

Neural networks are a set of algorithms modeled after the structure and function of the brain, specifically the way neurons interact with each other through synapses. Generally, neural networks consist of multiple layers of interconnected nodes, or artificial neurons. There are three main types of layers: input, hidden, and output.

As shown in Fig.1 Input neurons receive data, hidden neurons process that data and output neurons produce a result. The connections between neurons are determined by weights, which are adjusted during the learning process. The weights determine the strength of the connection between neurons and thus the importance of each neuron's input in producing the output.

There are also various types of neural networks, such as feedforward neural networks, recurrent neural networks, and convolutional neural networks, each with their structures and purposes.

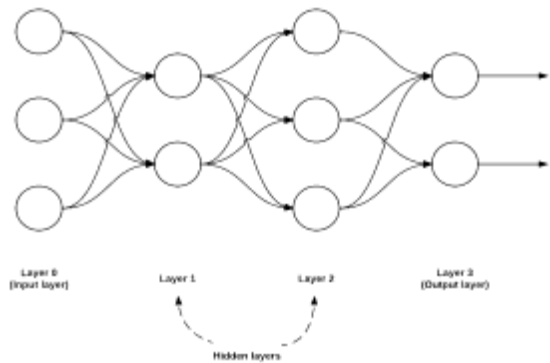

Fig.1 An example of a feedforward neural network with 3 input nodes, a hidden layer with 2 nodes, a second hidden layer with 3 nodes, and a final output layer with 2 nodes.

In this article, we are going to use the LeNet neural network as a neural network example and protect it against FGSM attacks. LeNet is a Convolutional Neural Network structure proposed by LeCun et al. (1998) [20]. LeNet refers to LeNet-5 and is a simple convolutional neural network. Convolutional Neural Networks are a kind of feed-forward neural network whose artificial neurons can respond to a part of the surrounding cells in the coverage range and perform well in large-scale image processing. It was one of the first successful neural networks for image recognition and classification tasks.

LeNet was originally designed for handwritten digit recognition, but it has since been used for a wide range of image recognition tasks, such as facial recognition, traffic sign recognition, and more.

The architecture of LeNet consists of seven layers, including two convolutional layers, two subsampling layers, and three fully connected layers. The convolutional layers are used to extract features from the input image, while the subsampling layers reduce the dimensionality of the feature maps. The fully connected layers are used for classification.

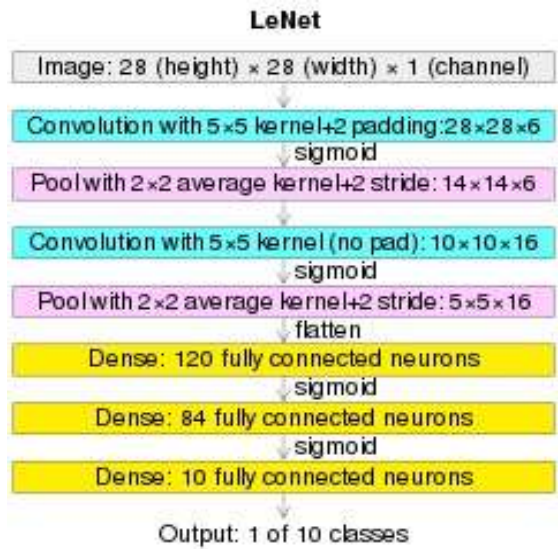

Fig.2 Convolution, pooling, dense layers of lenet model

As shown in Fig.2 (input image data with 32*32 pixels): LeCun et al. (1989) [21] showed that LeNet-5 consists of seven layers. In addition to input, every other layer can train parameters. In the figure, Cx represents the convolution layer, Sx represents the sub-sampling layer, Fx represents the complete connection layer and x represents the layer index.

Layer C1 is a convolution layer with six convolution kernels of 5x5 and the size of feature mapping is 28x28, which can prevent the information of the input image from falling out of the boundary of the convolution kernel.

Layer S2 is the subsampling/pooling layer that outputs 6 feature graphs of size 14x14. Each cell in each feature map is connected to 2x2 neighborhoods in the corresponding feature map in C1.

Layer C3 is a convolution layer with 16 5-5 convolution kernels. The input of the first six C3 feature maps is each continuous subset of the three feature maps in S2, the input of the next six feature maps comes from the input of the four continuous subsets, and the input of the next three feature maps comes from the four discontinuous subsets. Finally, the input for the last feature graph comes from all feature graphs of S2.

Layer S4 is similar to S2, with a size of 2x2 and an output of 16 5x5 feature graphs.

Layer C5 is a convolution layer with 120 convolution kernels of size 5x5. Each cell is connected to the 5*5 neighborhood on all 16 feature graphs of S4. Here, since the feature graph size of S4 is also 5x5, the output size of C5 is 1*1. So S4 and C5 are completely connected. C5 is labeled as a convolutional layer instead of a fully connected layer, because if LeNet-5 input becomes larger and its structure remains unchanged, its output size will be greater than 1x1, i.e. not a fully connected layer.

F6 layer is fully connected to C5, and 84 feature graphs are output [22].

## 3 Machine Unlearning

Machine unlearning is the process of modifying or removing the data used to train a machine learning model, to improve its accuracy or performance [23]. Machine learning models are typically trained on large datasets, but sometimes these datasets can include biased or irrelevant data that negatively impact the model's performance. In such cases, unlearning can help to remove the problematic data and retrain the model with more accurate or relevant data.

The emerging research area of Machine Unlearning (MU) is highlighted as a potential solution. MU, still in its infancy, seeks to remove specific data points from ML models, effectively making them 'forget' completely specific information. If successful, MU could provide a feasible means to manage the overabundance of information and ensure better protection of privacy and IP [24].

There are several methods for unlearning in machine learning, including data poisoning, data deletion, and data perturbation [25]. Data poisoning involves adding malicious or misleading data to the dataset to trick the model into making incorrect predictions. Data deletion involves simply removing the problematic data from the dataset. Data perturbation involves modifying the data in the dataset to remove any biases or irrelevant information.
Another important aspect of machine learning unlearning is privacy. When data is used to train a machine learning model, it can often contain sensitive or personal information about individuals. Therefore, it is important to ensure that the unlearning process is performed in a way that protects individuals' privacy and does not compromise their personal information.
Overall, machine learning unlearning is an important tool for improving the accuracy and performance of machine learning models, and it will likely become even more important as machine learning continues to advance and become more widely used in various industries.

## 4 Proposed Method

In this section, we will explain our method to protect the LeNet neural network against FGSM attacks using machine unlearning. In the first step, we train the LeNet neural network. After training and feeding the network, we calculate the accuracy of the model's performance. In the second step, we attack the model by FGSM and calculate the performance accuracy of the model after the attack. In the third step, we identify adversarial samples using the loss function and have them to the machine unlearning. In the fourth step, the machine unlearning removes the adversarial samples that have the most damage. Finally, we repeat the third and fourth steps until the model achieves optimal accuracy.

### 4.1 Lenet Neural Network Training

The training process for a LeNet neural network involves feeding the network with a set of labeled images as input and adjusting the weights of the network's layers through a series of backpropagation iterations to minimize the error between the network's predicted outputs and the true labels. Several frameworks can be used for implementing

LeNet and other neural network architectures, such as TensorFlow, PyTorch, and Keras. Each framework offers its own set of tools and APIs to facilitate the training process and enable the user to fine-tune the network's parameters and hyperparameters.

An implementation of LeNet neural network architecture in Python using Keras API:

```
import keras
from keras.models import Sequential
from keras.layers import Dense, Dropout
from keras.layers import Flatten, MaxPooling2D, Conv2D

def LeNet(input_shape, num_classes):
    model = Sequential()
    model.add(Conv2D(32, kernel_size=(3,3), activation='relu', input_shape=(28,28,1)) )
    model.add(Conv2D(64, kernel_size=(3,3), activation='relu'))
    model.add(MaxPooling2D(pool_size=(2,2)))
    model.add(Dropout(0.25))
    model.add(Flatten())
    model.add(Dense(128, activation='relu'))
    model.add(Dropout(0.5))
    model.add(Dense(n_classes, activation='softmax'))
return model
```

Fig.3 Lenet Neural Network Training

According to Fig.3 you can use this function to create a LeNet model with the desired input shape and the number of output classes. For example, to create a model for image classification with 10 classes on 28x28 grayscale images, you can do:

model = LeNet((28, 28, 1), 10)

## 4.2 FGSM Attack on Lenet

Fast Gradient Sign Method (FGSM) is a technique used in deep learning to craft adversarial examples, which are inputs to a model that are intentionally designed to deceive the model into producing incorrect output. This method involves adding a small perturbation to the input image in the direction of the gradient of the loss function, which causes the output of the model to change. FGSM is one of the most popular and effective adversarial attack techniques used in deep learning.

The code for FGSM can vary depending on the framework or library you are using, but The basic idea is to take the gradient of the loss function concerning the input image, and then perturb the image in the direction that maximizes the loss function. This perturbation is typically scaled by a small constant epsilon, which controls the strength of the attack.

An example implementation of the Fast Gradient Sign Method (FGSM) attack on LeNet using Keras:

```
import numpy as np
import tensorflow as tf
from tensorflow import keras
from tensorflow.keras import layers

# Define the LeNet neural network architecture
model = keras.Sequential(
    [
        layers.InputLayer(input_shape=(28, 28)),
        layers.Reshape(target_shape=(28, 28, 1)),
        layers.Conv2D(filters=6, kernel_size=(3, 3), activation='relu'),
        layers.AveragePooling2D(),
        layers.Conv2D(filters=16, kernel_size=(3, 3), activation='relu'),
        layers.AveragePooling2D(),
        layers.Flatten(),
        layers.Dense(units=120, activation='relu'),
        layers.Dense(units=84, activation='relu'),
        layers.Dense(units=10, activation='softmax'),
    ]
)

# Define a function to perform the FGSM attack
def fgsm_attack(model, x, epsilon):
    x_adv = tf.Variable(x.reshape((1, 28, 28, 1)))
    y_true = tf.constant(y_test[image_index].reshape((1, 10)), dtype=tf.float32)
    # Generate adversarial perturbation using FGSM
    with tf.GradientTape() as tape:
        predictions = model(inputs)
    loss = keras.losses.categorical_crossentropy(y_true, prediction)

    # Calculate gradient of loss with respect to input
    gradient = tape.gradient(loss, x_adv)

    # Calculate sign of gradient
    signed_grad = tf.sign(gradient)

    # Add perturbation to inputs
    x_adv.assign_add(epsilon * signed_grad)
    x_adv_array = x_adv.numpy()
    return x_adv_array

# Example usage
epsilon = 0.1
model = LeNet()

# Train model normally...
# Evaluate model normally...
# Generate perturbed examples using the FGSM attack
(x_train, y_train), (x_test, y_test) = keras.datasets.mnist.load_data()
x_train = x_train.astype('float32') / 255

x_test = x_test.astype('float32') / 255
y_train = keras.utils.to_categorical(y_train, 10)
y_test = keras.utils.to_categorical(y_test, 10)

perturbed_inputs = fgsm_attack(model, inputs, epsilon)
```

Fig.4 FGSM Attack on Lenet

In Fig.4 the FGSM_attack function applies the FGSM attack to a set of inputs. Note that this implementation assumes that the input images are grayscale with a height and width of 28 pixels and that the output is a vector of length 10 representing the log for each of the digits 0-9. You may need to modify this implementation work with your specific dataset and training/evaluation code.

### 4.3 Identify the adversarial examples

A Python code snippet that will generate adversarial examples using the FGSM attack and then Identify the examples that caused the highest loss:

```python
import tensorflow as tf
import numpy as np

# Load the LeNet model
model = tf.keras.models.load_model('lenet_model')

# Load the test data
(x_test, y_test), _ = tf.keras.datasets.mnist.load_data()

# Normalize the input data
x_test = x_test.astype('float32') / 255

# Predict the class probabilities for the adversarial examples
prediction = model.predict(x_adv)

# Calculate the loss for each adversarial example
y_true = tf.constant(y_test[image_index].reshape((1, 10)),
dtype=tf.float32)
loss = keras.losses.categorical_crossentropy(y_true, prediction)

# Generate adversarial examples using the FGSM attack
epsilon = 0.1
x_adv = x_test + epsilon * tf.sign(tf.gradients(model.output,
model.input)[0])

# Select the adversarial examples with the highest loss for
machine unlearning
selected_adv_examples = x_adv_array
```

Fig.5 Identify the adversarial examples

In Fig.5 this code assumes that you have already trained and saved a LeNet model for the MNIST dataset. It loads the test data and then generates adversarial examples using the FGSM attack with an epsilon value of 0.1. It then predicts the class probabilities for each adversarial example and calculates the loss.
Finally, it identifies the indices of the adversarial examples that caused the highest loss and selects those examples for machine unlearning.
Note that this is just one approach to identifying the most harmful adversarial examples for machine unlearning. Depending on your specific use case and model architecture, you may need to use a different approach to identify the most harmful examples.

### 4.4 Remove the selected adversarial examples

To remove the selected adversarial examples from the training dataset, you can use the following code:

```python
clean_train_dataset = []
clean_train_labels = []

for i in range(len(train_dataset)):
    if i not in selected_adv_examples:
        clean_train_dataset.append(train_dataset[i])
        clean_train_labels.append(train_labels[i])
```

Fig.6 Remove the selected adversarial examples

In Fig.6 Selected_adv_examples is a list of indices of the adversarial examples that you want to remove from the train_dataset. After removing the selected adversarial examples, you can retrain the LeNet model on the new dataset using the following code:

```python
model = LeNet()
model.compile(optimizer='adam', loss='categorical_crossentropy',
metrics=['accuracy'])
model.fit(np.array(clean_train_dataset),
np.array(clean_train_labels), batch_size=128, epochs=15,
validation_data=(test_dataset, test_labels))
```

Fig.7 Retrain the LeNet model

In Fig.7 Clean_train_dataset and clean_train_labels are the cleaned training dataset and labels, respectively. You can also use any other optimizer, loss function, or metrics based on your specific needs.

### 4.5 Selecting a new subset of examples for machine unlearning

Repeat the process by generating new adversarial examples using the updated model and then selecting a new subset of examples for machine unlearning.
A Python code snippet that you can generate new adversarial examples using the updated model and select a new subset of examples for machine unlearning:

```python
# Generate new adversarial examples using the updated model
test_adv_examples = fgsm_attack(model, test_dataset,
epsilon=0.1)

# Evaluate the new adversarial examples and select a subset for
machine unlearning
evaluated_adv_examples = evaluate_examples(model,
test_adv_examples, test_labels)
selected_adv_examples =
select_examples_for_unlearning(evaluated_adv_examples,
num_examples=100)

# Remove the selected adversarial examples from the training
dataset and retrain the model
clean_train_dataset = []
clean_train_labels = []

for i in range(len(train_dataset)):
    if i not in selected_adv_examples:
        clean_train_dataset.append(train_dataset[i])
        clean_train_labels.append(train_labels[i])

model.fit(np.array(clean_train_dataset),
np.array(clean_train_labels), batch_size=128, epochs=15,
validation_data=(test_dataset, test_labels))
```

Fig.8 Selecting a new subset of examples for machine unlearning

In Fig.8 generate_adversarial_examples is a function that generates adversarial examples using the updated model, evaluate_examples is a function that evaluates the adversarial examples and returns a list of tuples of the form (example, true_label, predicted_label), and select_examples_for_unlearning is a function that selects a subset of adversarial examples for machine unlearning based on some criteria (e.g., prediction confidence). After selecting the adversarial examples for unlearning, you can remove them from the training dataset and retrain the model on the cleaned dataset.

### 4.6 The desired level of the model robustness

An example of how you can continue the process of machine unlearning and adversarial example generation until the model achieves a desired level of robustness against the FGSM attack:

```
# Train the model on the original training dataset
model = LeNet()
model.compile(optimizer='adam', loss='categorical_crossentropy',
metrics=['accuracy'])
model.fit(train_dataset, train_labels, batch_size=128, epochs=15,
validation_data=(test_dataset, test_labels))

# Evaluate the performance of the trained model on both the
original test dataset and adversarial test examples
accuracy = model.evaluate(test_dataset, test_labels)[1]
accuracy_adv = model.evaluate(fgsm_attack(model, test_dataset,
epsilon=0.1), test_labels)[1]
print('Accuracy on original test dataset:', accuracy)
print('Accuracy on adversarial test dataset:', accuracy_adv)

# Continue unlearning process until desired level of robustness is
achieved
while accuracy_adv < 0.9:  # Set desired level of robustness to
90%
# Generate new adversarial examples and select subset for
unlearning
    adv_examples_for_unlearning = fgsm_attack(model,
test_dataset, epsilon=0.1)
    evaluated_adv_examples = evaluate_examples(model,
test_adv_examples, test_labels)
    selected_adv_examples =
adv_examples_for_unlearning(evaluated_adv_examples,
num_examples=100)

# Remove selected adversarial examples and retrain model
    clean_train_dataset = []
    clean_train_labels = []

    for i in range(len(train_dataset)):
      if i not in selected_adv_examples:
        clean_train_dataset.append(train_dataset[i])
        clean_train_labels.append(train_labels[i])

    model.fit(np.array(clean_train_dataset),
np.array(clean_train_labels), batch_size=128, epochs=15,
validation_data=(test_dataset, test_labels))

# Evaluate performance of updated model on both original and
adversarial test datasets
    accuracy = model.evaluate(test_dataset, test_labels)[1]
    accuracy_adv = model.evaluate(fgsm_attack(model,
test_dataset, epsilon=0.1), test_labels)[1]
    print('Accuracy on original test dataset:', accuracy)
    print('Accuracy on adversarial test dataset:', accuracy_adv)
```

Fig.9 The desired level of the model robustness

According to Fig.9 we train the model on the original training dataset, evaluate its performance on both the original and adversarial test datasets, and then continue the unlearning process until the model achieves a desired level of robustness (in this case, 90% accuracy on the adversarial test dataset). We evaluate the performance of the updated model after each round of unlearning to track its progress toward the desired level of robustness.

### 4.7 Evaluation results

You can evaluate the performance of the trained model on both the original test dataset and a set of adversarial test examples generated using the FGSM attack:

Table 1: Accuracy and loss function results of neural network model
For three cases: trained model, attacked model, and protected model

| Model | Accuracy | Loss |
|---|---|---|
| **Original test dataset** | 0.9932 | 0.0271 |
| **Adversarial test datasets** | 0.0796 | 2.3045 |
| **Updated model** | 0.9669 | 0.1546 |

- # Evaluate the performance of the trained model on the original test dataset
  loss, accuracy = model.evaluate(test_dataset, test_labels)
  print('Accuracy on original test dataset:', accuracy)

As shown in Table 1, after training the model on the original test dataset, the following accuracy has been obtained:
Accuracy on original test dataset:
0.9932000041007996

- # Generate adversarial examples using the FGSM attack
  test_adv_dataset = fgsm_attack(model, test_dataset, epsilon=0.1)

  # Evaluate the performance of the trained model on the adversarial test dataset
  loss_adv, accuracy_adv = model.evaluate(test_adv_dataset, test_labels)
  print('Accuracy on adversarial test dataset:', accuracy_adv)

After training on the model on the adversarial test dataset, the following accuracy has been obtained:
Accuracy on adversarial test dataset:
0.07959999889135361

- # Evaluate the performance of updated model adversarial test datasets
  accuracy_adv = model.evaluate(fgsm_attack(model, test_dataset, epsilon=0.1), test_labels)[1]
  print('Accuracy on adversarial test dataset:', accuracy_adv)

Accuracy of the updated model on adversarial test dataset:
0.9668999910354614

You can repeat this evaluation process multiple times as you continue to train the model and select examples for unlearning until you achieve the desired level of robustness against the FGSM attack.

This paper takes a unique approach called 'unlearning' instead of 'retraining.' In this project, we're challenging traditional methods by exploring how we can achieve remarkable results by unlearning existing patterns and biases. By doing so, we're breaking free from limitations and opening up new possibilities. It's an innovative approach that allows us to redefine the boundaries and uncover fresh insights.

While conducting our research on Protecting the Lenet against FGSM Attack Using Machine Unlearning, we extensively searched for projects and articles specifically addressing this research area. Unfortunately, we did not come across a project or article that directly aligns with our work. However, it is worth noting that there have been notable studies in the broader field of adversarial attacks and machine learning. Papers such as [19], and [26] have delved into various adversarial attack techniques, including the FGSM attack, and their impact on machine learning models.

## 5 Conclusions

This paper explores the use of machine unlearning to enhance the robustness of the LeNet neural network against the Fast Gradient Sign Method (FGSM) attack. The paper proposes a novel method for implementing machine unlearning, which is a process of removing selected data samples from a model's training data to prevent overfitting and increase the model's generalization ability. The authors apply this technique to the LeNet neural network and show that it can significantly improve the network's resistance to FGSM attacks. The paper presents experimental results that demonstrate the effectiveness of the proposed method, showing that the LeNet network with machine unlearning performs better than the original network in terms of accuracy and robustness against FGSM attacks. Overall, this paper provides a valuable contribution to the field of machine learning security and offers a promising solution for enhancing the security of neural networks against adversarial attacks.

## Declarations of Conflict of Interest

The authors declared that they have no conflicts of interest to this work.

## Ethical approval

This article does not contain any studies with human participants or animals performed by any of the authors.

## Research Data Policy and Data Availability Statements

The datasets generated during and/or analyzed during the current study are available from the corresponding author on reasonable request.

## Author contributions

Jahanian instigated this work, led the research, supervised the student projects, and led the data analysis and writing.
Khorasani contributed to the implementation of AI models, data collection and analysis in this work.
Rastgarpour contributed to the supervision of the student projects and led the technical writing in this paper.